\documentclass{article}

\usepackage{arxiv}

\usepackage{amsmath}
\usepackage{multirow}

\usepackage[utf8]{inputenc} 
\usepackage[T1]{fontenc}    
\usepackage{hyperref}       
\usepackage{url}            
\usepackage{booktabs}       
\usepackage{amsfonts}       
\usepackage{nicefrac}       
\usepackage{microtype}      
\usepackage{cleveref}       
\usepackage{lipsum}         
\usepackage{graphicx}
\usepackage[numbers]{natbib}
\usepackage{doi}

\title{Designing Dynamic Pricing for Bike-sharing Systems via Differentiable Agent-based Simulation}

\date{}

\newif\ifuniqueAffiliation
\uniqueAffiliationtrue

\ifuniqueAffiliation 
\author{ \href{https://orcid.org/0009-0007-1195-866X}{\includegraphics[scale=0.06]{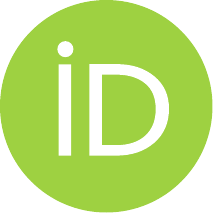}\hspace{1mm}Tatsuya Mitomi}\\
	Fujitsu Limited \\
    Kawasaki \\
    Japan \\
	\texttt{mitomi.tatsuya@fujitsu.com} \\
	\And
	\href{https://orcid.org/0000-0001-9247-4104}{\includegraphics[scale=0.06]{orcid.pdf}\hspace{1mm}Fumiyasu Makinoshima} \\
	Fujitsu Limited \\
    Kawasaki \\
    Japan \\
	\texttt{f.makinoshima@fujitsu.com} \\
	\And
	{\hspace{1mm}Fumiya Makihara} \\
	Fujitsu Limited \\
    Kawasaki \\
    Japan \\
	\texttt{makihara.fumiya@fujitsu.com} \\
	\And
	\href{https://orcid.org/0009-0003-8985-241X}{\includegraphics[scale=0.06]{orcid.pdf}\hspace{1mm}Eigo Segawa} \\
	Fujitsu Limited \\
    Kawasaki \\
    Japan \\
	\texttt{segawa.eigo@fujitsu.com} \\
}

\renewcommand{\shorttitle}{}

\hypersetup{
pdftitle={Designing Dynamic Pricing for Bike-sharing Systems via Differentiable Agent-based Simulation},
pdfsubject={cs.MA, cs.LG},
pdfauthor={T. Mitomi et al.},
pdfkeywords={Bike-sharing systems, Differentiable agent-based simulation, Differentiable programming, Dynamic pricing},
}

\begin{document}
\maketitle

\begin{abstract}
Bike-sharing systems are emerging in various cities as a new ecofriendly transportation system.
In these systems, spatiotemporally varying user demands lead to imbalanced inventory at bicycle stations, resulting in additional relocation costs. Therefore, it is essential to manage user demand through optimal dynamic pricing for the system. 
However, optimal pricing design for such a system is challenging because the system involves users with diverse backgrounds and their probabilistic choices. 
To address this problem, we develop a differentiable agent-based simulation to rapidly design dynamic pricing in bike-sharing systems, achieving balanced bicycle inventory despite spatiotemporally heterogeneous trips and probabilistic user decisions.
We first validate our approach against conventional methods through numerical experiments involving 25 bicycle stations and five time slots, yielding 100 parameters.
Compared to the conventional methods, our approach obtains a more accurate solution with a 73\% to 78\% reduction in loss while achieving more than a 100-fold increase in convergence speed.
We further validate our approach on a large-scale urban bike-sharing system scenario involving 289 bicycle stations, resulting in a total of 1156 parameters.
Through simulations using the obtained pricing policies, we confirm that these policies can naturally induce balanced inventory without any manual relocation.
Additionally, we find that the cost of discounts to induce the balanced inventory can be minimized by setting appropriate initial conditions.
\end{abstract}


\section{Introduction}

\begin{NoHyper}
\footnotetext{This work has been submitted to the IEEE for possible publication. Copyright may be transferred without notice, after which this version may no longer be accessible.}
\end{NoHyper}

Dynamically designing service pricing is crucial for ensuring business profitability and sustainability in various industries, such as online marketplaces~\cite{Schlosser}, energy grids~\cite{Bandyopadhyay}, and transportation~\cite{Saharan}.
In transportation, bike-sharing systems (BSS), introduced in various cities as a new ecofriendly transportation system~\cite{DAlmeida}, provide an example in which designing optimal prices is crucial for sustainable business continuity.
A challenge within this system is the significant imbalance in bicycle inventory across stations due to the spatiotemporally varying trip demands of users~\cite{DellAmico,Vallez}.
This imbalance forces system operators to manually relocate bicycles from stations with excess inventory to stations lacking bicycles (i.e., operator-based relocation) to enhance users convenience by minimizing system unavailability~\cite{Li}. 
However, operator-based relocation disrupts the sustainability of this ecofriendly system, negatively affecting both revenue and reliability~\cite{Nath}.
To address this issue, a concept known as user-based relocation has recently gained attention~\cite{Shui,Thomas}.
Unlike operator-based relocation, user-based relocation balances inventories by offering discounts to encourage users to return bicycles to stations experiencing shortages, rather than relying on manual operator intervention.

Optimal pricing design is crucial for enabling user-based relocation in BSS.
Common approaches to designing such pricing strategies rely on mathematical optimization or bandit algorithms to induce user participation~\cite{Yuan,Singla}.
These approaches model aggregate user demand and choice behavior, facilitating efficient problem-solving with theoretical guarantees.
However, they simplify user behavior to mathematically formulate problems by overlooking user heterogeneity.
When designing optimal pricing in BSS, considering individual heterogeneity is crucial, as diverse user preferences induce different responses to pricing, resulting in probabilistic behaviors.
Therefore, mathematical modelling approaches that ignore such heterogeneity create a significant gap between human behavior and mathematical models, questioning the effectiveness of pricing strategies in real-world applications.

\begin{figure}[t]
    \centering
    \includegraphics[scale = 0.28]{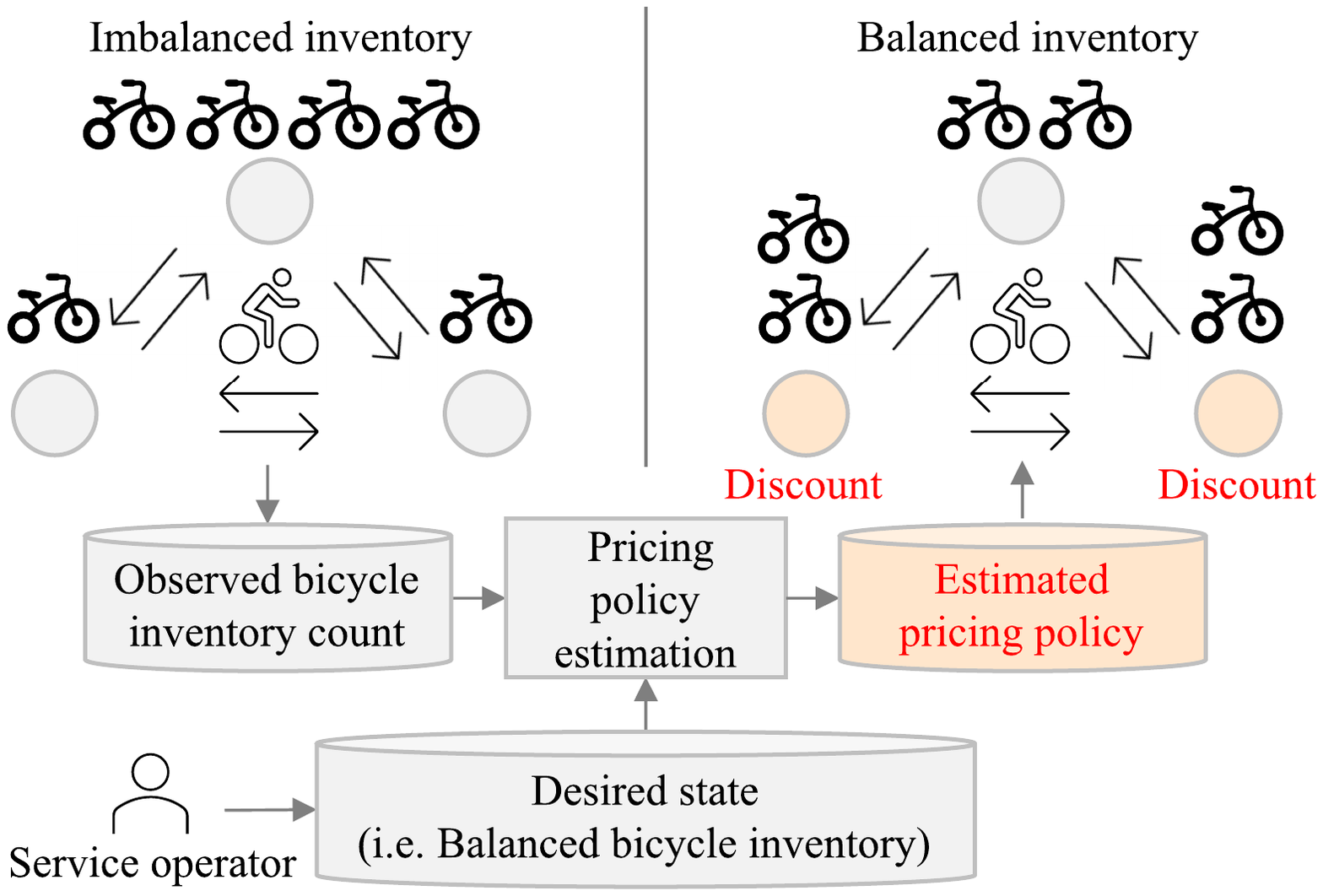}
    \vspace{-0.4cm}
    \caption{Overview of a dynamic pricing method for user-based relocation in BSS.}
    \label{fig:overroll_fig}
\end{figure}

Agent-based models (ABMs) offer a robust solution to this issue as they naturally accounts for user heterogeneity and probabilistic choices through interacting agents~\cite{Bonabeau,Fagnant,Lu}.
This approach facilitates the exploration of optimal pricing designs to encourage user-based relocation while accounting for user heterogeneity and probabilistic choices.
However, ABMs cannot be expressed analytically and involve stochastic components due to probabilistic decision-making, making them non-differentiable and typically requiring gradient-free optimization algorithms, such as genetic algorithms, for parameter optimization ~\cite{Calvez}.
Therefore, optimization using ABMs requires multiple simulation runs to optimize pricing parameters, which becomes computationally expensive, particularly when the number of parameters is large.
Consequently, conventional approaches encounter difficulty when solving real-world problems, which typically have numerous optimizing parameters.
In a BSS problem, the number of parameters to be optimized becomes exponentially large with the number of bicycle stations and the frequency of pricing modifications.
This requires a more computationally efficient optimization approach for solving real-world problems using ABMs.

To address this challenge, we introduce a differentiable ABM for designing the pricing of BSS. 
This approach rapidly provides solutions to the optimization problem, even with stochastic components and numerous optimizing parameters.
Differentiable ABMs are a new simulation technique that use differentiable programming for automatic differentiation (AD)~\cite{Griewank,Margossian}, enabling efficient and precise calculation of gradients.
By embedding differentiable programming into ABMs, we construct end-to-end differentiable simulators that facilitate efficient gradient-based optimization---capabilities not feasible with conventional, nondifferentiable ABMs~\cite{Andelfinger,Chopra,Dyer}.
In this study, we apply this simulation technique to the pricing problem of BSS to induce user-based relocations (Fig. ~\ref{fig:overroll_fig}).
Although differentiable ABMs have been applied to various problems, such as epidemiology~\cite{Chopra} and finance~\cite{Dyer}, they are yet to be used in dynamic pricing contexts.
Thus, to the best of our knowledge, this study is the first to formulate dynamic pricing problems using differentiable ABMs.
Our contributions can be summarized as follows:
\begin{itemize}
    \item We formulate the dynamic pricing problem in BSS using a differentiable ABM. 
    \item In a numerical experiment involving 100 pricing parameters, we demonstrate that our approach can derive a pricing policy that can naturally induce balanced inventory without any operator-based relocations.
    This policy achieves a 73\% to 78\% reduction in loss and over 100 times faster convergence speed than that of conventional methods.
    \item We conduct large-scale BSS deployment setting experiments involving 1156 pricing parameters and confirm that our approach can derive a pricing policy that can naturally induce balanced inventory without any operator-based relocations at this scale. 
    With our approach, the number of simulation runs required for optimization remains unchanged, even when the number of optimized parameters increases 10-fold, from 100 to 1156.
    \item We show that the cost of inducing these user-based relocations can be minimized by imposing appropriate initial pricing parameters for the estimation process.
\end{itemize}

\section{Related Works}
\subsection{Mathematical Optimization Approaches for Designing Dynamic Pricing}
Mathematical optimization is widely used for designing dynamic pricing strategies across industries.
In the transportation sector, several studies have focused on designing dynamic pricing in BSS by mathematically modeling user demand~\cite{Yuan,Haider}.
These studies mathematically model user demands and choices, facilitating the use of efficient optimization algorithms that provide theoretical guarantees.
A similar mathematical modeling approach has been applied to dynamic pricing in the retail industry~\cite{Shimizu}, where the relationship between product price and user demand is captured.
As demonstrated in these examples, mathematical optimization can be used to design dynamic pricing in various fields; however, it does not account for the heterogeneity in individual responses to pricing in its modeling assumptions.
When individuals purchase services or products, the heterogeneity of their responses to pricing diversifies the impact of price on decision-making, leading to probabilistic behaviors. 
Therefore, failing to address these probabilistic behaviors when modeling individual purchasing behaviors may diminish the effectiveness of the designed dynamic pricing strategy. 

As a time-varying modeling approach, methods based on Markov decision processes (MDPs) and bandit algorithms have been widely studied~\cite{Puterman,Lattimore}. 
Several studies have explored designing dynamic pricing based on MDP formulations considering examples of BSS and ride-sharing services ~\cite{Pan,MChen}.
Additionally, research on designing dynamic pricing using bandit algorithms has been reported in other transportation and retail fields~\cite{Singla,Jia,Mao}. 
However, these studies face challenges in describing the probabilistic behaviors of individuals, as deterministic models fail to account for the varying impact of price on decision-making.

\subsection{ABM-based Optimization Approaches}
ABM is a modeling approach that incorporates the heterogeneity of users with probabilistic choices. 
Optimizations using ABM typically employ gradient-free methods.
The grid search method, a conventional gradient-free method, has been used in several studies~\cite{Chang,Romero-Brufau}.
Genetic algorithms are also used as an advanced, gradient-free optimization method in ABM~\cite{Calvez}.
These methods explore the parameter space using gradient-free update rules to obtain an optimal solution.
When optimizing a large number of parameters, as is often encountered in real-world problems, these methods typically require numerous simulation runs and take a long time to converge to a solution.

Finite differences (FD) can be used to approximate the parameter gradients in ABM~\cite{Andelfinger}, enabling the use of gradient-based optimization algorithms through these approximated gradients.
However, this method requires a large number of simulation runs as the number of optimization parameters increases because it approximates  parameter gradients using the results from multiple simulations with slightly varied parameters. 
Consequently, the computational time constraints hinder ABM scalability when many parameters require optimization.

Surrogate modeling is a method that approximates ABM with other models~\cite{Barton,Bhosekar}, typically employing differentiable models such as neural networks.
Surrogate modeling allows for the approximate calculation of gradients of parameters in ABM, enabling the use of gradient-based optimization algorithms when implemented with differentiable programming.
However, individual-level interactions in the original ABM are not expressed in surrogate modeling\cite{Anirudh,Andelfinger}. 
Consequently, methods that accurately represent the probabilistic behaviors of individuals while enabling efficient gradient-based optimization have not yet been proposed for designing dynamic pricing.
Our approach enables gradient-based optimization in ABM that represents the probabilistic behaviors of individuals. 

\section{Simulation Settings}

\begin{figure}[t] 
    \centering
    \includegraphics[scale = 0.4]{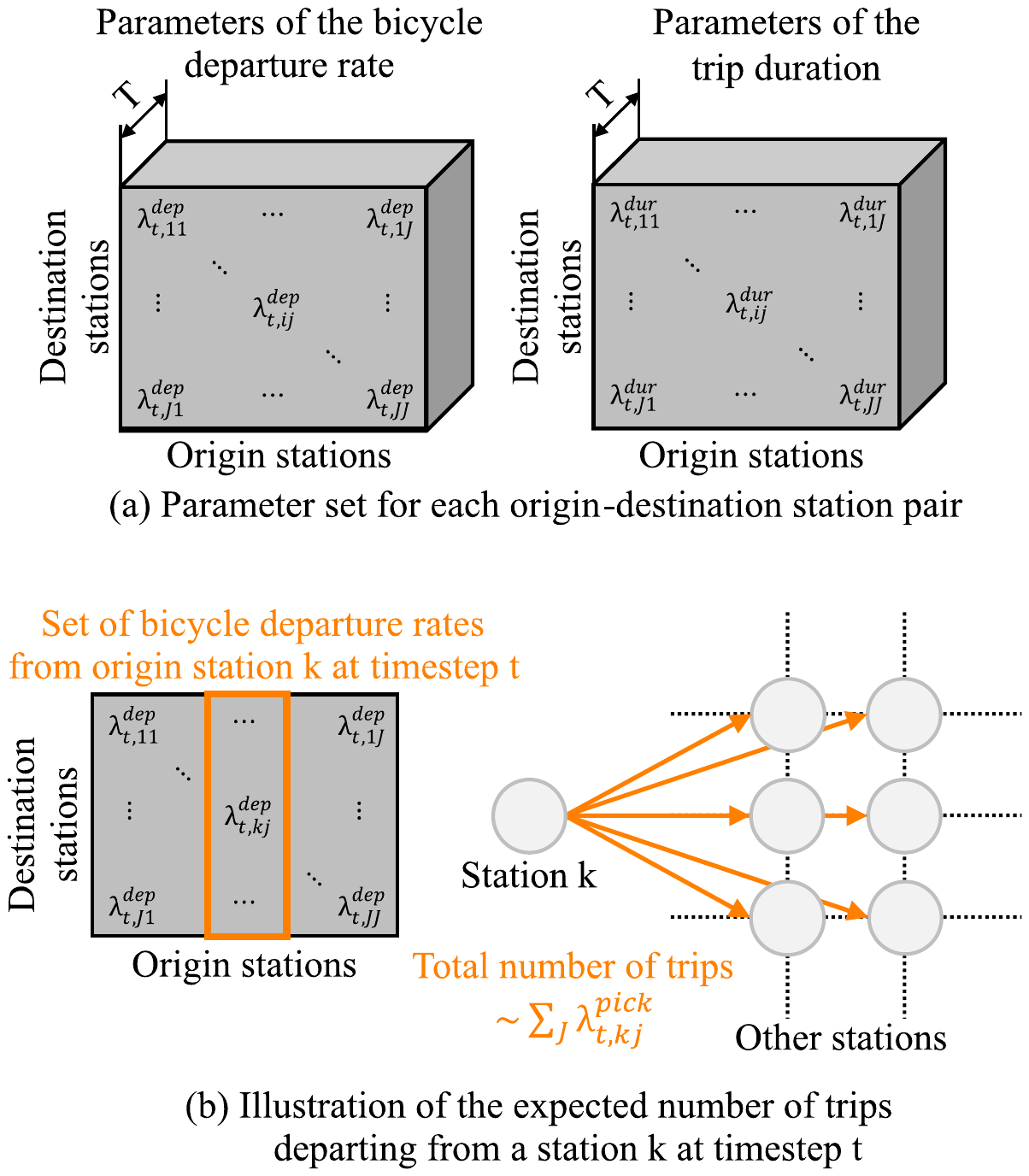}
    \vspace{-1.8cm}
    \caption{Representation of trip demand (a) Parameter set for each origin-destination station pair. (b) Illustration of the expected number of trips departing from a given station $k$ at timestep $t$. The trip demand is set between each station; therefore, the expected number of trips departing from a specific station $k$ is the sum of the expected trips to other stations}
    \label{fig:simulation}
\end{figure}

\subsection{Assumptions}
In our ABM formulation, users act as agents who rent bicycles from an origin station and travel to a destination station, which may include round trips to the same station.
These trips are simulated over $T$ time steps between $J$ stations. 
Each origin-destination pair is associated with a trip demand that describes the potential number of agents traveling between them.
Agents are generated based on this demand and may individually revise their destination choices. 
These decisions are influenced by the discount $p_{t,i}$ at other stations and the distance from their originally intended station to other stations.
Under these assumptions, our ABM primarily explores the discounts $p_{t,i}$ that can achieve a balanced bicycle inventory without any other intervention.

\subsection{Representation of Trip Demand}
Trip demands is represented using probabilistic distributions that closely approximate actual trip demands in BSS modeling~\cite{Thomas,Christine}.
Specifically, trip demand is modeled using two types of probabilistic distributions: the Poisson distribution for representing the bicycle departure rate (i.e., the timing when the user starts using the bicycle) and the exponential distribution for representing trip duration. 
These distributions determine the number of potential agents traveling from an origin station $i$ to a destination station $j$ at timestep $t$, with a trip duration of $dur_{t, ij}$. 
These are characterized by a parameter set for each origin--destination station pair $(i, j)$ at the given timestep $t$.
Thus, our ABM has two sets of $J \times J \times T$ parameters for each distribution representing trip demand, as illustrated in Fig~\ref{fig:simulation}a.
These parameters are described as $\lambda^{dep} _{t, ij}$ and $\lambda^{dur} _{t, ij}$ for the distribution of the bicycle departure rate and trip duration, respectively. 
According to the definition of the Poisson distribution, $\lambda^{dep}_{t, ij}$ represents the expected value of the number of bicycle departures, corresponding to the number of trips departing from station $i$ to station $j$ at timestep $t$. 
Therefore, the expected value of the number of trips departing from a certain station $k$ at timestep $t$ is given by $\sum_{j} \lambda^{dep}_{t, kj}$, as illustrated in Fig.~\ref{fig:simulation}b.

\subsection{Modeling the Probabilistic Choices of Users}
\begin{figure}[t] 
    \centering
    \includegraphics[scale = 0.4]{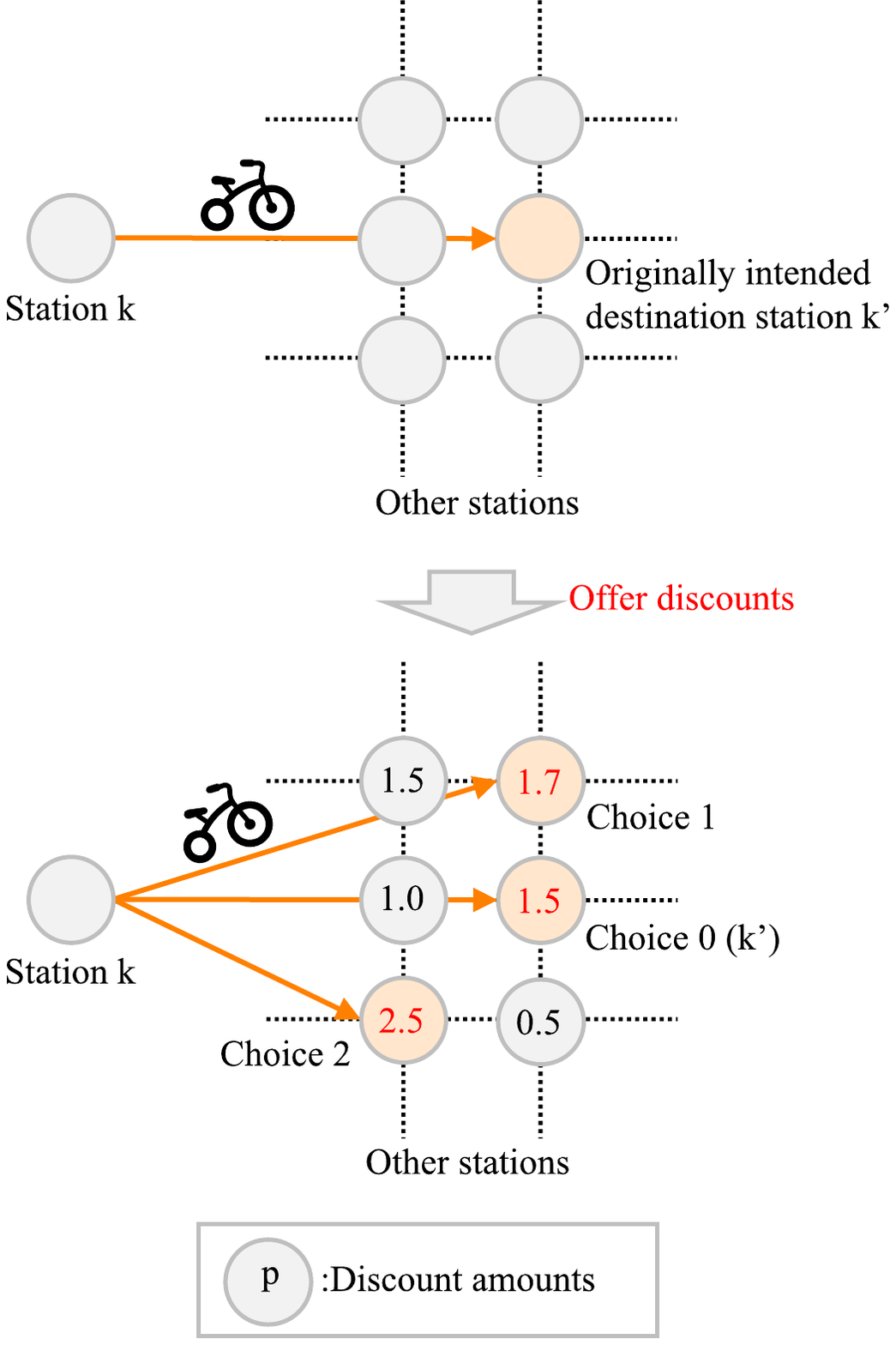}
    \vspace{-0.5cm}
    \caption{Illustration of an example choice set of stations. In this example, choices 1 and 2 are alternative choices to the originally intended destination station $k'$ (choice 0) because the discount amounts for both choices are higher than those for $k'$. The attributes of the utilities for each station are as follows: choice 1 has $\varDelta p_{t,s} = 0.2$ and $d = 1$, choice 2 has $\varDelta p_{t,s} = 1.0$ and $d = 2$, and choice 0 (the originally intended station) has $\varDelta p_{t,s} = 0$ and $d = 0$.}
    \label{fig:alternative_stations}
\end{figure}

We describe changes in destination station based on the probabilistic choices of agents.
Typically, there is no motivation to change the destination station from the originally intended one, which is determined by the bicycle departure rate distribution.
However, if another station offers a higher discount than the originally intended one, it can serve as motivation for changing the destination station. 
Therefore, changing the destination station from the originally intended one does not occur without such motivation.
To model this, we define a choice set of stations that includes the originally intended destination station and those offering higher discounts than the originally intended one, as illustrated in Fig~\ref{fig:alternative_stations}.
Therefore, the choice set of stations for the originally intended destination station $j$ is described by
\begin{equation}
    S_{t, j} = \{ s \in J \mid s = j \lor p_{t, s} > p_{t, j}\},
    \label{eq:candidate_set_of_return_bays}
\end{equation}
where $S _{t, j}$ is the choice set of stations that the agents consider choosing at timestep $t$, and $s$ denotes a station within the choice set $S _{t, j}$.

Our ABM uses a discrete choice model (DCM) to simulate agents' probabilistic choice of return station.
DCMs are widely adopted for modeling individual choices from discrete choice alternatives and are commonly employed in ABMs to represent the probabilistic choices of agents~\cite{Train,Tzouras}.
In DCM, decision-makers are assumed to make choices by evaluating the utility of each choice alternative, such as selecting a destination station in our ABM.
We assume two attributes as the utilities: the difference in discounts and the distance between the originally intended destination station $j$ and the station $s$ in the choice set, represented by $\varDelta p_{t, s}$ and $d_{j, s}$, respectively. 
Therefore, the utility of station $s$ at timestep $t$ is given by
\begin{equation}
    u_{t, s} = w_{discount} \cdot \varDelta p_{t, s} + w_{distance} \cdot d_{j, s} + asc_{s},
    \label{eq:discount_bay_utility}
\end{equation}
where $u_{t, s}$ is the utility of station $s$ in the choice set $S _{t, j}$ at timestep $t$.
$w_{discount}$ and $w_{distance}$ represent the weights associated with the differences in discount and distance, respectively.
$asc_{s}$ (i.e., the alternative specific constant) represents the inherent attractiveness to station $s$ in the choice set $S _{t, j}$.
Selecting a station other than the originally intended one incurs a "switching" burden.
To reflect this, the $asc_{s}$ for nonintended stations (i.e., choices involving "switching" destination stations) is set lower than that of the originally intended station.
Using this utility, the probability of selecting station $s$ in the choice set $S _{t, j}$ is given by
\begin{equation}
    P_{t}(s) = \frac {exp(u_{t, s})} {\sum_{k \in S_{t, j}} exp( u_{t, k})},
    \label{eq:discount_bay_probability}
\end{equation}
where $P_{t}(s)$ is the probability of selecting station $s$ in the choice set $S _{t, j}$ at timestep $t$.

\subsection{Formulation of Differentiable ABM}
Agent choices and trip demands are probabilistic models described by discrete distributions, which typically result in ABM being nondifferentiable.
Therefore, we introduce some approximating methods that serve as a sampling method for these probabilistic models, making our ABM differentiable through AD.
The Gumbel-Softmax reparameterization~\cite{Jang,Maddison} is a method for continuously approximating categorical samples from discrete probabilistic distributions. 
It enables the calculation of gradients via AD even when discrete probabilistic processes are involved.
Using this approach, sampling from the discrete probability distribution $P _{i}$ can be expressed as follows:
\begin{equation}
    y_{i} = \frac {exp( (\log {P_{i}} + g_{i}) / \tau)}  {\sum_{k=1}^{K} exp( (\log {P_{k}} + g_{k}) / \tau)},
    \label{eq:gumbel_softmax}
\end{equation}
where $K$ represents the number of categories in the discrete probabilistic distribution, $y_{i}$ denotes $K$-dimensional sample vectors, $g_{i}$ represents samples drawn from i.i.d. Gumbel(0, 1) distribution, and $\tau$ represents the softmax temperature parameter.
Equation~\eqref{eq:gumbel_softmax} approximates sampling from a discrete probabilistic distribution by using samples from a continuous Gumbel distribution, enabling differentiation of the sampling process.
The DCM models the agents' choices using finite discrete probabilistic distributions; hence, the sampling from~\eqref{eq:discount_bay_probability} can be replaced with~\eqref{eq:gumbel_softmax}, making the process differentiable.

The Poisson and exponential distributions, applied to model bicycle departure rates and trip durations, are infinite probabilistic distributions. 
Generalized Gumbel-Softmax (GenGS)~\cite{Joo} is employed as a sampling method for these infinite discrete probabilistic distributions.
GenGS sets an integer limit on the number of discrete categories in an infinitely discrete probabilistic distribution, reducing an infinite set of categories to a finite set. 
This approach allows for reparameterization with Gumbel-Softmax even for infinite discrete probabilistic distributions.
Let $n$ denotes the upper limit on the number of categories in an infinite discrete probabilistic distribution. 
The probability of each discrete category $i$ in this infinite discrete probabilistic distribution is approximated as 
\begin{equation}
    P_{i} = 
    \begin{cases} 
        P_{i}                    & \text{if } i < n \\
        1 - \sum_{k=1}^{n-1} P_{k} & \text{if } i = n \\
        0                        & \text{if } i > n.
    \end{cases}
    \label{eq:GenGS}
\end{equation}
After approximating a finite discrete distribution using~\eqref{eq:GenGS}, sampling is performed according to~\eqref{eq:gumbel_softmax}.
This process allows the infinite discrete sampling to be approximated using differentiable sampling. 
These approximation methods make our ABM end-to-end differentiable, enabling efficient gradient-based optimization using AD.

\subsection{Bicycle Inventory in Each Station}
The simulated number of bicycle inventories at station $i$ at time step $t$ is denoted by $\hat{I}_{t,i}$. 
Initially, $\hat{I}_{t,j}$ is set to the number of bicycles at time step 0 and updates based on the trips of the agents at every time step $t$. 
At timestep $t$, $\hat{I}_{t,i}$ is updated based on the number of departure and return events at each station $i$.

\section{Experiments}

\begin{figure*}[t] 
    \centering
    \includegraphics[scale = 1]{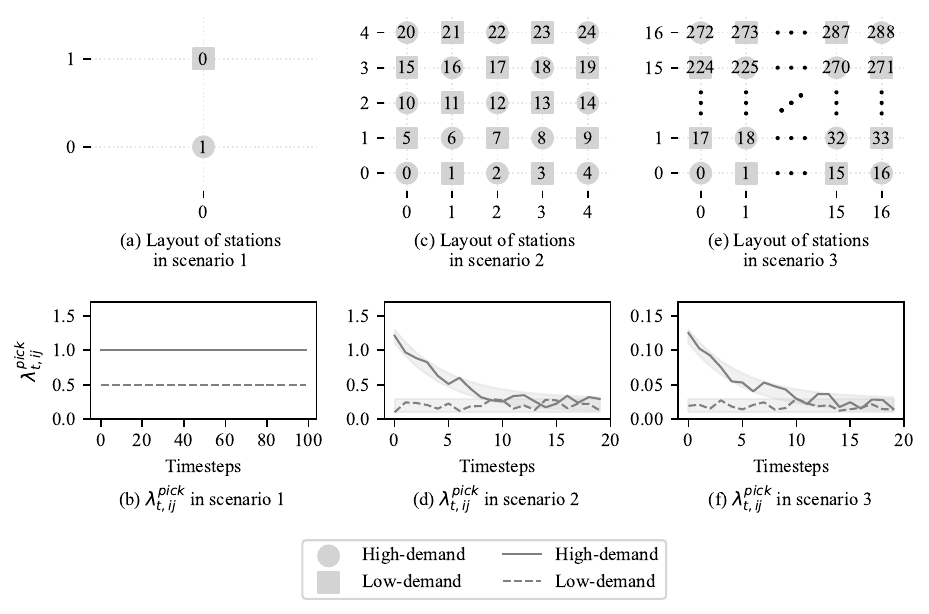}
    \vspace{-0.0cm}
    \caption{Scenarios 1, 2, and 3 for numerical experiments: (a) Layout of stations in Scenario 1. 
    The distance between the two stations is 1.
    (b) Changes of $\lambda^{dep}_{t, ij}$ in Scenario 1.
    (c) Layout of stations in Scenario 2. The stations are arranged in a grid pattern, with each side having a length of 1.
    Because trips are conducted along the grid, the distance between each station is described by the Manhattan distance. 
    (d) Changes of $\lambda^{dep}_{t, ij}$ in Scenario 2. 
    (e) Layout of stations in Scenario 3. Arrangement of stations and the distance between each station are the same as in Scenario 2.
    (f) Changes of $\lambda^{dep}_{t, ij}$ in Scenario 3.
    }
    \label{fig:experiments_setting}
\end{figure*}

\subsection{Experimental Setup}
\subsubsection{Scenario Settings}
\paragraph{Scenario 1: Simplified setting.} To verify the fundamental characteristics of our method, we generate a simple synthetic trip demand scenario involving two stations: one designed as the high-demand station and the other as the low-demand station (Fig. \ref{fig:experiments_setting}a).
The demand are time-invariant, with $\lambda^{dep} _{t,01}=1.0$ and $\lambda^{dep} _{t,10}=0.5$ (Fig. \ref{fig:experiments_setting}b), and $\lambda^{dur} _{t,ij}=1$.
Similarly, the discount parameter $p _{t,j}$ is set to be time-invariant.
Thus, only two parameters must be estimated: $p _{0}$ and $p _{1}$.  
Concerning parameters related to the DCM, the weights associated with the differences in discount and distance are set to $w_{\text{discount}} = 1$ and $w_{\text{distance}} = -1$, respectively. 
Here, $asc_{s}$ is set to $0$ if station $s$ is the originally intended destination station, as determined by a bicycle departure rate distribution. 
Conversely, $asc_{s}$ is set to $-1$ if station $s$ is not the originally intended one.
The total number of simulation time steps $T$ is set to 100, with each station starting with an initial inventory of 100 bicycles. 

\begin{figure}[t]
    \centering
    \includegraphics[scale=1]{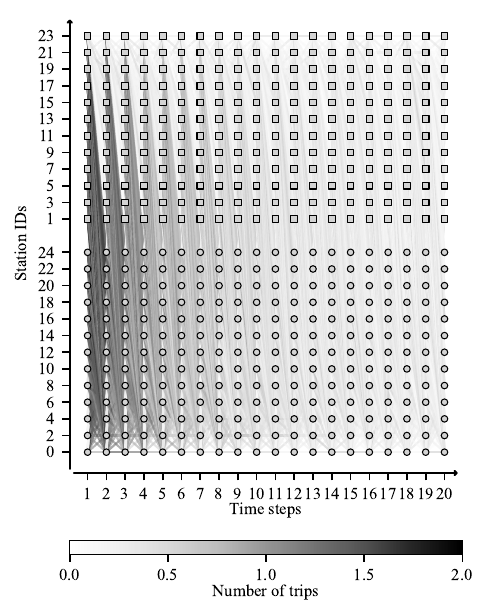}
    \vspace{-0.3cm}
    \caption{Simulated number of trips between stations each time step without discounts under the $demand _{test}$ in Scenario 2 (averaged over 30 simulations). The solid lines represent the number of trips.}
    \label{fig:original_result}
\end{figure}

\paragraph{Scenario 2: Small-scale setting.} In real-world scenarios within limited spaces such as university campuses, the deployment of BSS involves dozens of stations~\cite{Thomas}.
As a scenario for deploying BSS at this scale, we assume a simulation period of 10 hours with 0.5 hour time steps (i.e., a total simulation time of $T=20$) across 25 stations, which include both high- and low-demand stations (Fig. \ref{fig:experiments_setting}c), with an initial bicycle inventory of 100.
In this scenario, $p_{t,j}$ remains constant for each set of five time steps.
Thus, the total number of discount parameters to be estimated is 100 $(25 \text{ stations} \times (20 \text{ time steps} / 5 \text{ steps}))$.
It is considered that trips over longer distances take more time.
To reflect this, the trip duration distribution parameter is designed as $\lambda^{dur} _{t,ij} = 5/d_{ij}$.
$\lambda^{dep} _{t,ij}$ is designed as a time-variant parameter: for high-demand stations, $\lambda^{dep}_{t, ij} = \exp(-0.2t)+0.1+U(0, 0.2)$, and for low-demand stations, $\lambda^{dep}_{t, ij} = 0.1+U(0, 0.2)$ (Fig. ~\ref{fig:experiments_setting}d).
$U(a, b)$ denotes a uniform distribution between $a$ and $b$.
The parameters related to DCM are set to be consistent with those in Scenario 1.

\paragraph{Scenario 3: Large-scale setting.} In a large-scale urban deployment of BSS, the number of stations can exceed 100~\cite{Li}. 
To verify our method in large-scale deployments of BSS, we conduct simulations across 289 stations, including both high- and low-demand stations (Fig. \ref{fig:experiments_setting}e). 
The simulation period, time step, $p_{t,j}$ adjustment, and initial bicycle inventory are consistent with those in Scenario 2.
Consequently, the total estimated number of discount parameters is 1156 $(289 \text{ stations} \times (20 \text{ time steps} / 5 \text{ steps}))$.
As illustrated in Fig.~\ref{fig:simulation}b, the expected value of the number of trips departing from each station is proportional to both the number of stations and the parameter $\lambda^{dep}_{t,ij}$.
Given that the number of stations is approximately 10 times larger than in Scenario 2, the parameter $\lambda^{dep}_{t,ij}$ is set to be 10 times smaller than in Scenario 2.
Specifically, for high-demand stations, $\lambda^{dep} _{t,ij}$ is calculated as $\frac{1}{10}(\exp(-0.2t) + 0.1 + U(0, 0.2))$, and for low-demand stations, it is $\frac{1}{10}(0.1 + U(0, 0.2))$ (Fig.~\ref{fig:experiments_setting}f).
This ensures that the number of trips beginning from each station per time step closely matches Scenario 2, enabling experiments under identical inventory conditions.
The parameters related to DCM are set to be consistent with those in Scenarios 1 and 2.

\begin{figure}[t]
    \centering
    \includegraphics[scale=1]{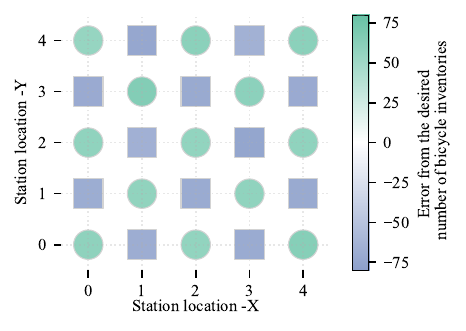}
    \vspace{-0.3cm}
    \caption{Simulated number of bicycle inventories at the final time step without discounts under the $demand _{test}$ in Scenario 2 (averaged over 30 simulations). These numbers are shown as a heatmap of the error from the desired number of bicycle inventories $I_{T, j}$, with the station location pattern aligned with Fig. \ref{fig:experiments_setting}c.}
    \label{fig:original_result_inventory}
\end{figure}

\subsubsection{Evaluation}
\paragraph{Scenario 1: Simplified setting.} In this scenario, we evaluate whether the estimation of discount parameters can be performed using AD and examine the effect of varying the initial discount parameters on the estimations.
As the aim of the estimation is to obtain discount parameters that induce user-based relocation without any additional intervention, we set the mean squared error of the bicycle inventory at the final time step as a loss function, which represents whether the discount parameters achieve a balanced inventory.
The ground-truth discount parameters are set as $p_{0}=0.5$ and $p_{1}=2.5$.
Using these parameters, the desired bicycle inventory at the final time step ${I} _{T, j}$ is generated from a single simulation run.

\begin{figure}[t]
    \centering
    \includegraphics[scale=1]{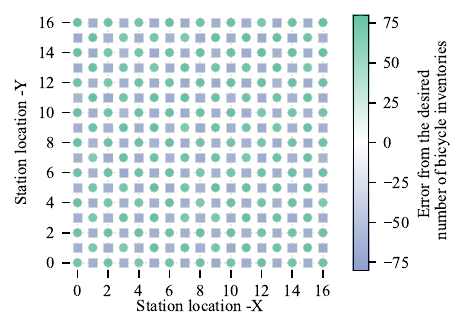}
    \vspace{-0.2cm}
    \caption{Simulated number of bicycle inventories at the final time step without discounts under the $demand _{test}$ in Scenario 3 (averaged over 30 simulations). These numbers are shown as a heatmap of the error from the desired number of bicycle inventories $I_{T, j}$, with the station location pattern aligned with Fig. \ref{fig:experiments_setting}e.}
    \label{fig:largescale_result_inventory}
\end{figure}

\begin{figure}[t]
    \centering
    \includegraphics[scale = 1]{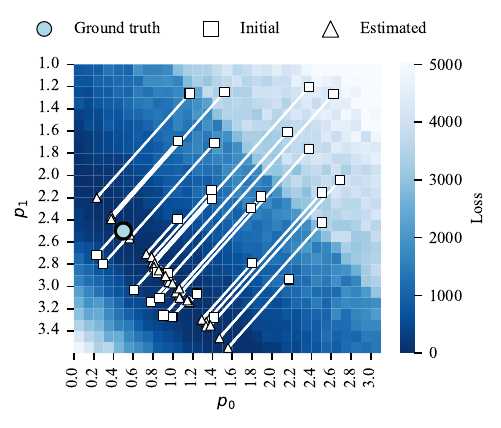}
    \vspace{-0.5cm}
    \caption{Heatmap of the loss and the trajectory of the estimation in Scenario 1.}
    \label{fig:loss_surface}
\end{figure}

\paragraph{Scenario 2: Small-scale setting.} In this scenario, we evaluate three key aspects: (1) whether the estimated discount parameters can induce user-based relocation without any additional intervention, even when applied to unknown trip demands; (2) whether the convergence speed of the proposed method is faster than that of conventional methods; and (3) how the implementation of user-based relocation using estimated discount parameters affects the operator costs. 
To conduct these evaluations, we generate two trip demands using different random seeds. 
One set ($demand _{est}$) is used to estimate the discount parameters, while the other ($demand _{test}$) is used to test the performance of the estimated discount parameters.
For evaluation, a simulation is repeated 30 times by applying the estimated discounts to agents generated from $demand _{test}$, which is not used in the discount parameter estimation. Subsequently, the average number of trips and bicycle inventory are assessed.
Without discounts, the number of trips between each station at each time step and the bicycle inventory at the final time step under the $demand _{test}$ are imbalanced, as shown in Fig. \ref{fig:original_result} and \ref{fig:original_result_inventory}, respectively.
Similar to Scenario 1, we set the mean squared error of bicycle inventory at the final time step as the loss function. 
To observe the effect of varying the initial discount parameters, we prepare two types of initial discount parameter patterns. Initial pattern 1 is generated by randomly selecting values from a uniform distribution U(0, 0.1), while initial pattern 2 is generated from a uniform distribution U(1, 1.1).
In reality, the desired number ${I} _{T, j}$ is a perfect state requested by the operator and is not generated by the simulator; thus, it might be unachievable by the simulator. 
Considering this, ${I} _{T, j}$ is not generated by simulation but is manually set.
We set ${I} _{T, j}$ to 90, assuming that some bicycles remain in transit and are not returned to the stations in the final time step. 

To compare convergence speed, we estimate the discount parameters using three methods, including two conventional methods, with the same trip demand data.
(1) Differential evolution (DE), a gradient-free parameter optimization method, has proven effective in many experiments~\cite{Calvez}.
(2) Gradient descent with gradient estimation using FD.
(3) Differentiable ABM (our method).

In addition to comparing convergence speed, we assess the performance of the estimated parameters and the operator costs associated with these estimated parameters.
Regarding the operator cost incurred from implementing user-based relocation using estimated discount parameters, we define the operator cost as the total discount amount across all time steps and stations.
\begin{equation}
    Cost = \sum_{t=1}^T \sum_{j=1}^J p_{t,j},
\end{equation}
where $Cost$ indicates the operator cost of user-based relocation using discounts.

\paragraph{Scenario 3: Large-scale setting.} In this scenario, we evaluate the scalability of our method to large-scale urban deployments of BSS.
Similar to Scenario 2, we generate two sets of trip demands using different random seeds. 
One set ($demand _{est}$) is used to estimate the discount parameters, while the other ($demand _{test}$) is used to evaluate the performance of the estimated discount parameters, with ${I} _{T, j}$ set to 80. 
For evaluation, a simulation is repeated 30 times by applying the estimated discounts to agents generated from $demand _{test}$, which is not used in the discount parameter estimation. Subsequently, the average bicycle inventory is assessed.
Without discounts, the number of bicycles in the inventory at the final time step under the $demand _{test}$ is imbalanced, as shown in Fig. \ref{fig:largescale_result_inventory}.
As in Scenarios 1 and 2, we use the mean squared error of the bicycle inventory at the final time step as the loss function.
When handling hundreds of stations, as in this scenario, conventional methods such as DE and FD cannot estimate discount parameters within a realistic timeframe, making it infeasible to compare their speed and performance with the proposed method.
Consequently, Scenario 3 is not utilized as a benchmark for evaluating these conventional methods.

\begin{figure}[t]
    \centering
    \includegraphics[scale=1]{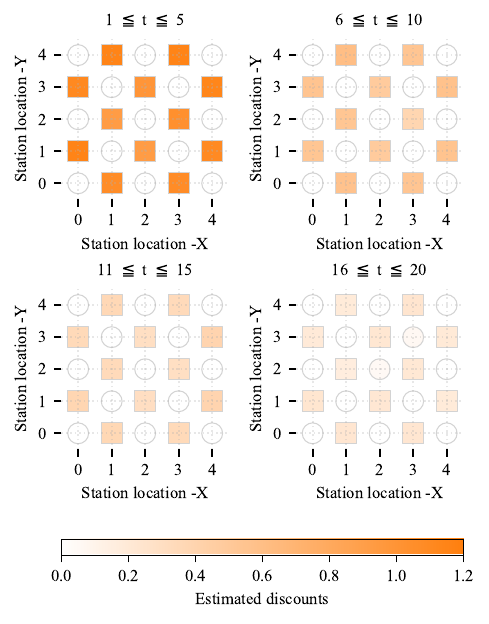}
    \vspace{-0.4cm}
    \caption{Estimated discount parameters for each station at each time step in Scenario 2. These parameters are shown as a heatmap, with the station location pattern aligned with Fig. \ref{fig:experiments_setting}c.}
    \label{fig:discounted_result_discounts}
\end{figure} 

\begin{figure}[t]
    \centering
    \includegraphics[scale=1]{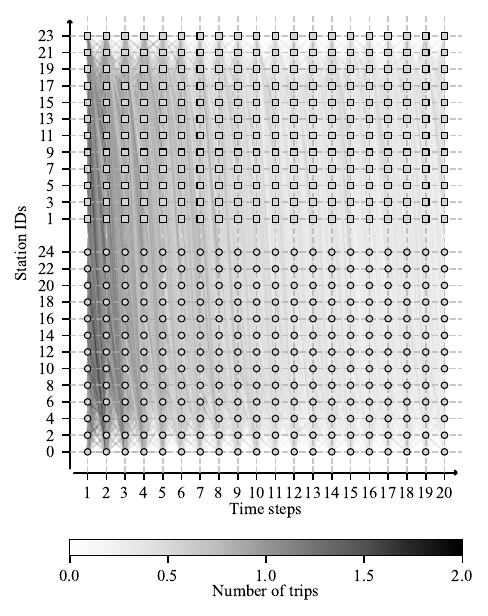}
    \vspace{-0.4cm}
    \caption{Simulated number of trips between stations each time step with estimated discounts under the $demand _{test}$ in Scenario 2 (averaged over 30 simulations). The solid lines represent the number of trips.}
    \label{fig:discounted_result}
\end{figure} 

\begin{figure}[t]
    \centering
    \includegraphics[scale=1]{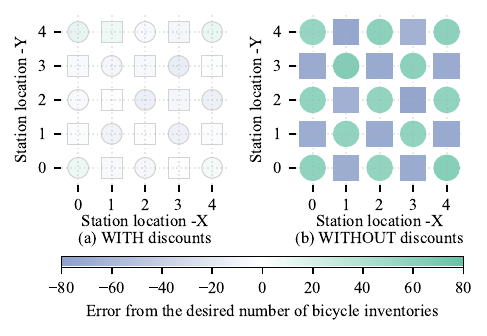}
    \vspace{-0.4cm}
    \caption{Simulated number of bicycle inventories at the final time step under the $demand _{test}$ in Scenario 2 (averaged over 30 simulations). These numbers are shown as a heatmap of the error from the desired number of bicycle inventories $I_{T, j}$, with the station location pattern aligned with Fig. \ref{fig:experiments_setting}c. (a) Simulated number with estimated discounts. (b) Simulated number without discounts (re-displayed in Fig. \ref{fig:original_result_inventory}).}
    \label{fig:discounted_result_inventory}
\end{figure}

\subsubsection{Implementation Details}
For DE, we use the implementation of SciPy~\cite{Virtanen}, a widely used scientific computing library, with a population size of 100, mutation rate of 0.5, and recombination rate of 0.7.
For FD, both gradient estimation via FD and the gradient descent algorithm are implemented using operations of PyTorch~\cite{Paszke}, a widely used machine learning framework, with the learning rate set to $10^{-5}$ and the step size for FD set to 0.1.
In our method, implementation AD and the SGD optimizer of PyTorch are used, with the learning rate set to $10^{-3}$ in Scenarios 1 and 2 and to $10^{-2}$ in Scenario 3.
The softmax temperature $\tau$ is set to 1.0. 
We conduct five batches of simulation runs for these methods. 
The following squared error of the bicycle inventory at the final time step is implemented as a loss function:
\begin{equation}
    Loss = \frac{1}{J} \sum_{j=1}^{J} \left( I _{T, j} - \hat{I} _{T, j} \right)^2,
    \label{eq:loss}
\end{equation}
where $Loss$ represents the loss function, and ${I} _{T, j}$ and $\hat{I} _{T, j}$ represent the desired and simulated number of bicycle inventories at station $j$ at the final time step $T$, respectively.
All experiments are performed on a CPU with 2$\times$Xeon Gold 5218 CPU (2.3 GHz $\times$14 core) and 192 GB of RAM.

\begin{figure}[t]
    \centering
    {\includegraphics[scale=1.0]{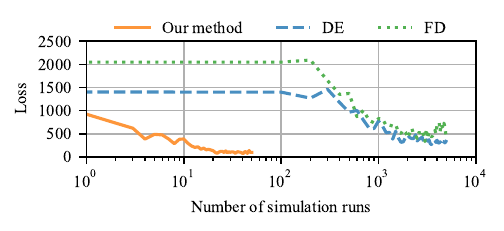}}
    \vspace{-0.3cm}
    \caption{Relationship between the number of simulation runs and the changes in the loss in Scenario 2.}
    \label{fig:loss_histories}
\end{figure}

\begin{table}[t]
    \centering
    \caption{\textbf{Operator cost and final loss for each initial discount parameter pattern in Scenario 2.}}
    \renewcommand{\arraystretch}{1.2} 
    \begin{tabular}{p{65pt}p{50pt}p{50pt}p{50pt}p{50pt}}
        \hline
        \multirow{2}{*}{\textbf{Methods}} & \multicolumn{2}{c}{\textbf{Initial pattern 1}} & \multicolumn{2}{c}{\textbf{Initial pattern 2}} \\
        & \textbf{Cost} & \textbf{Loss} & \textbf{Cost} & \textbf{Loss} \\
        \hline
        Our method & 24.80 & 96.84 & 105.5 & 63.62 \\
        DE & 89.88 & 363.7 & 105.5 & 1227 \\
        FD & 69.10 & 441.3 & 98.44 & 345.7 \\
        \hline
    \multicolumn{5}{p{300pt}}{Initial pattern 1 is generated by randomly selecting the initial values of discount parameters for estimation from a uniform distribution $U(0, 0.1)$, while initial pattern 2 is generated from a uniform distribution $U(1, 1.1)$}
    \end{tabular}
    \label{tab:estimated total discounts and the loss}
\end{table}

\subsection{Experimental Results}

\subsubsection{Scenario 1: Simplified Setting}
Fig. \ref{fig:loss_surface} shows a heatmap of the loss in simulations for each set of discount parameters $p_{j}$, along with the trajectory of estimations from 30 initial $p_{j}$ values. 
The heatmap indicates that the optimal solution is not a single point but rather a line. 
Consequently, as our method employs gradient-based optimization, the estimations of $p_{j}$ converge to the multiple solutions closest to the initial value along the loss landscape.
In the Scenario 1 setting, only agents heading towards stations with smaller discounts consider whether to change their destination. 
According to~\eqref{eq:discount_bay_utility}, when $p_{0} - p_{1}$ is constant, the utility for the station remains unchanged, and the probability of changing destinations stays constant.
Therefore, when $p_{0} - p_{1}$ matches the ground truth, the probability of changing destinations also aligns with the ground truth, resulting in the bicycle inventory at the final time step matching the ground truth as well. 
Consequently, any parameter set where $p_{0} - p_{1}$ is identical to the ground truth is an optimal solution.
The results presented in Fig. \ref{fig:loss_surface} are consistent with this discussion. 
Simultaneously, this suggests that the possibility of controlling the discount parameter pattern obtained solely through the selection of initial values, which could lead to the control of the cost of user-based relocation.
Based on this, we hypothesize that choosing initial discount parameters close to zero will cause the discount parameters to converge near zero, reducing operator cost of user-based relocation while maintaining the same loss. 
In Scenario 2, we further examine this strategy for initial parameter setting.

\subsubsection{Scenario 2: Small-scale Setting}  
\paragraph{Evaluation of the estimated discount parameters.} Fig.~\ref{fig:discounted_result_discounts} displays the estimated discount parameters derived from initial discount parameter pattern 1, using $demand _{est}$ in Scenario 2. 
It is shown that the estimated discount parameters are time-variant across all time steps, along with time-varying demand $\lambda^{dep}_{t, ij}$.
Evaluation simulations are performed by applying these estimated discounts to agents generated from $demand _{test}$.
Fig.\ref{fig:discounted_result} shows the number of simulated trips at each timestep of these evaluation simulations.
As shown in Fig.\ref{fig:discounted_result}, the estimated discount parameters help to equalize the number of trips between stations compared to the number of trips without applying discounts (Fig. \ref{fig:original_result}).
Fig.\ref{fig:discounted_result_inventory}a shows the simulated bicycle inventory at the final timestep of these evaluation simulations.
As shown in Fig. \ref{fig:discounted_result_inventory}a, the bicycle inventory at the final time step, after applying the estimated discount parameters to the agents generated from $demand _{test}$, is balanced.
These results confirm that our method can estimate discount parameters that achieve user-based relocation without additional interventions.
        
\begin{figure}[t]
    \centering
    {\includegraphics[scale=1]{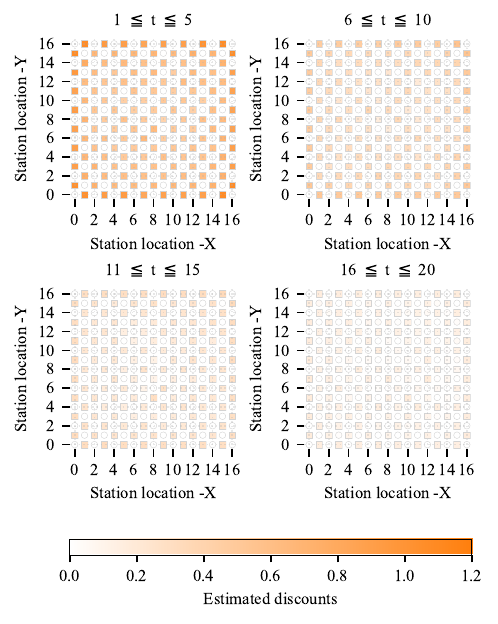}}
    \vspace{-0.3cm}
    \caption{Estimated discount parameters for each station at each time step in Scenario 3. These parameters are shown as a heatmap, with the station location pattern aligned with Fig. \ref{fig:experiments_setting}e.}
    \label{fig:largescare_result_discounts}
\end{figure}

\begin{figure}[t]
    \centering
    {\includegraphics[scale=1]{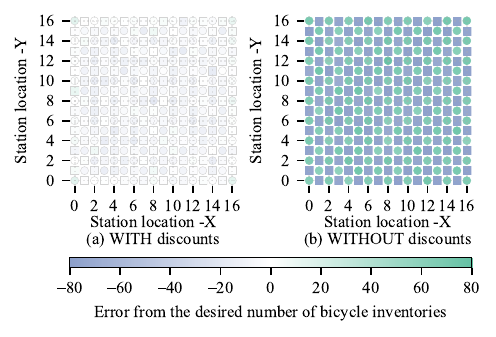}}
    \vspace{-0.3cm}
    \caption{Simulated number of bicycle inventories at the final time step under the $demand _{test}$ in Scenario 3 (averaged over 30 simulations). These numbers are shown as a heatmap of the error from the desired number of bicycle inventories $I_{T, j}$, with the station location pattern aligned with Fig. \ref{fig:experiments_setting}e. (a) The number with estimated discounts.(b) The number without discounts (re-displayed in Fig. \ref{fig:largescale_result_inventory}).}
    \label{fig:largescale_result_inventory_discounts}
\end{figure}

\begin{figure}[t]
    \centering
    {\includegraphics[scale=1]{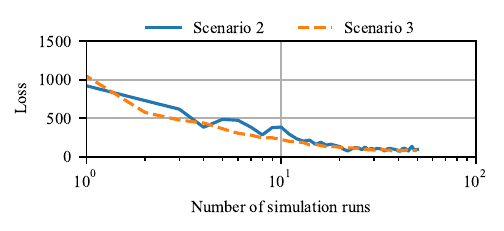}}
    \vspace{-0.3cm}
    \caption{Relationship between the number of simulation runs and the changes in loss observed in our method for Scenarios 2 and 3.}
    \label{fig:compare_senario23}
\end{figure}

\paragraph{Comparison of the convergence speed with existing methods.} In this comparison, we verify the difference in the number of simulation runs as the difference in estimation speed, as the calculation time of the parameter updates is negligible compared with the simulation runtime.
Fig.~\ref{fig:loss_histories} illustrates the relationship between the number of simulation runs and the changes in the loss with initial discount parameters pattern 1.
Regarding DE and FD, although the estimation is conducted over a period that is 100 times longer than that of our method, neither approach converges to the level of loss achieved by our method. 
Here, our method achieves a loss value that is 73\% better than that of DE and 78\% better than that of FD while achieving a 100 times faster convergence speed.
Unlike our approach, DE explores the parameter space using a gradient-free update rule, which may result in parameter update directions that are not optimal because it does not use gradients.
Conversely, FD updates the parameters using a gradient descent algorithm similar to our method but requires a number of simulations proportional to the number of parameters for each gradient estimation.
These factors explain why both methods take more than 100 times longer to converge than our method.
Regarding DE, the efficiency of the updates could potentially be improved by increasing the population size.
However, increasing population size also increases the number of simulations required for each update. 
Even with the current population size of 100, the number of simulations per update of 100 exceeds the total number of simulations required for our method to converge by 50, making further increases in convergence speed through this adjustment impractical.

\paragraph{Verification of the operator cost and the selection of initial discount parameters.} 
Table \ref{tab:estimated total discounts and the loss} lists the $cost$ and loss for each initial discount parameter pattern obtained by each algorithm.
As hypothesized in Scenario 1, the results demonstrate that the discount parameters estimated using initial value pattern 1 yield an 76\% lower $cost$ than pattern 2, while maintaining comparable loss values.
This reveals that setting the initial values for estimating the discount parameters to near zero can reduce the cost associated with user-based relocation using the estimated discount parameters.
Although FD also achieves a relatively small $cost$, the loss values do not reach the same level as our method, indicating that it does not obtain a price that can balance the number of bicycle inventories at the final time step.

\subsubsection{Scenario 3: Large-scale Setting}
Fig.~\ref{fig:largescare_result_discounts} presents the estimated discount parameters obtained using $demand _{est}$ in Scenario 3. 
Similar to the result in Scenario 2, it is shown that the estimated discount parameters are time-variant across all time steps, along with time-varying demand $\lambda^{dep}_{t, ij}$.
Fig.~\ref{fig:largescale_result_inventory_discounts}a illustrates the bicycle inventory at the final timestep from the simulation, where the estimated discount parameters are applied to agents generated from $demand _{test}$.
As shown in Fig.~\ref{fig:largescale_result_inventory_discounts}a, the bicycle inventory at the final time step is balanced.
These results confirm that our method can estimate discount parameters that achieve user-based relocation without additional interventions even in large-scale urban networks of BSS that include 289 stations with 1156 parameters.

Fig.~\ref{fig:compare_senario23} illustrates the relationship between the number of simulation runs and changes in loss compared to Scenario 2.
Although the estimated number of parameters is more than 10 times greater than in Scenario 2, the estimation for Scenario 3 converges to closely match the loss value with an identical number of simulation runs in Scenario 2.
This is because our method estimates the gradients of each parameter using AD, enabling the gradients of all parameters to be calculated simultaneously in a single simulation run.
Therefore, even if the number of parameters to be estimated increases, our method can estimate the discount parameters without requiring an increased number of simulation runs.

\section{Discussion}
We propose a differentiable ABM to address the dynamic pricing problem, with a focus on user-based relocation. 
By using synthetic data with 100 pricing parameters, we demonstrate that our differentiable ABM approach can develop a dynamic pricing policy that naturally achieves balanced inventory without operator-based relocations. 
This approach leads to a 73–-78\% reduction in loss and achieves convergence over 100 times faster than conventional methods.
Moreover, we demonstrate that our differentiable ABM approach can be applied to large-scale networks of BSS that include 289 stations with 1156 pricing parameters.
We also show that the costs for operators in BSS related to user-based relocations can be minimized by simply selecting appropriate initial pricing parameters during the estimation process.

The computational speed required for estimating optimal dynamic pricing presents significant challenges, particularly when considering individual probabilistic choices using ABM.
This is because ABM typically features nondifferentiable probabilistic components, often requiring gradient-free optimization algorithms, such as genetic algorithms~\cite{Calvez}.
This study demonstrates that incorporating automatic differentiation into ABM overcomes this challenge by significantly accelerating the computation speed for estimating optimal solutions in dynamic pricing problems in BSS. 
Similar dynamic pricing challenges exist in BSS and various fields, such as online marketplaces \cite{Schlosser}, energy grids \cite{Bandyopadhyay}, and transportation \cite{Saharan}.
In these fields, considering individual heterogeneity is crucial for designing optimal pricing, as diverse user preferences induce varied responses to pricing, resulting in probabilistic behaviors.
Modeling user behavior in these fields fundamentally mirrors our BSS modeling, as it involves user purchase demand (similar to trip demand in our ABM) and behavioral changes owing to pricing (analogous to destination station selection in our ABM).
Therefore, although we currently focus on the dynamic pricing problem in BSS, our results suggest that incorporating automatic differentiation into ABM can be applied to other dynamic pricing problems, given the similarities across these settings.

Future work will involve validating our method against real-world data.
Our method uses simplified human behavior models specifically a discrete choice model with a linear utility function. 
These linear utility functions reproduce human behavioral tendencies~\cite{Train}, suggesting that our current method can derive effective pricing even with real-world data.
However, owing to the simplicity of these functions, they may not fully express the complex behaviors occurring in the real world. 
In such a case, leveraging recent data-driven methods, including deep learning models, could enable the replication of complex behaviors and facilitate the determination of effective pricing within our framework.
These methods offer significant advantages in representing complex behaviors through nonlinear utility functions in discrete choice models~\cite{Sifringer, Makinoshima}.
Because such methods also use AD for training, they can be integrated into our differentiable ABM, thereby enhancing human behavior modeling while maintaining differentiability throughout the process.
This enhanced differentiable ABM will facilitate real-time dynamic pricing, ultimately ensuring business profitability and sustainability across various industries and society.

\bibliographystyle{unsrtnat}

\end{document}